\documentclass[10pt,twocolumn,letterpaper]{article}

\usepackage[pagenumbers]{iccv} 

\usepackage{tcolorbox}

\newcommand{\model}{\textit{MAmmoTH-VL2}\xspace}
\newcommand{\data}{\textsc{VisualWebInstruct}\xspace}
\newcommand{\best}[1]{\textbf{#1}}
\newcommand{\secondbest}[1]{\underline{#1}}
\definecolor{LightCyan}{rgb}{0.88,1,1}

\definecolor{iccvblue}{rgb}{0.21,0.49,0.74}
\usepackage[pagebackref,breaklinks,colorlinks,allcolors=iccvblue]{hyperref}
\usepackage{svg}
\def\paperID{11334} 
\def\confName{ICCV}
\def\confYear{2025}

\title{VisualWebInstruct: Scaling up Multimodal Instruction Data \\ through Web Search}

\author{$^{k,t}$Yiming Jia\thanks{Published during an internship at University of Waterloo}~, $^n$Jiachen Li, $^m$Xiang Yue, $^z$Bo Li, $^x$Ping Nie, $^y$Kai Zou, $^k$Wenhu Chen\\
$^k$University of Waterloo, 
$^t$University of Toronto,
\\
$^n$University of California, Santa Barbara,
$^m$Carnegie Mellon University,\\
$^z$Nanyang Technological University,
$^x$Independent, 
$^y$Netmind.ai\\
\{yiming.jia@mail.utoronto.ca, wenhuchen@uwaterloo.ca\}
\\
\url{https://tiger-ai-lab.github.io/VisualWebInstruct}
}

\begin{document}
\maketitle

\begin{abstract}
Vision-Language Models have made significant progress on many perception-focused tasks. However, their progress on reasoning-focused tasks remains limited due to the lack of high-quality and diverse training data. In this work, we aim to address the scarcity of reasoning-focused multimodal datasets. We propose VisualWebInstruct, a novel approach that leverages search engines to create a diverse and high-quality dataset spanning multiple disciplines, including mathematics, physics, finance, and chemistry, etc. Starting with a meticulously selected set of 30,000 seed images, we employ Google Image Search to identify websites containing similar images. We collect and process HTML data from over 700K unique URLs. Through a pipeline of content extraction, filtering, and synthesis, we construct a dataset of approximately 900K question-answer (QA) pairs, with 40\% consisting of visual QA pairs and the remaining comprising text-based QA pairs. Models fine-tuned on VisualWebInstruct demonstrate significant performance improvements: (1) fine-tuning on Llava-OV results in 10-20 absolute points improvement across benchmarks, and (2) fine-tuning from MAmmoTH-VL yields a 5 absolute points gain across benchmarks. Our best model, MAmmoTH-VL2, achieves state-of-the-art performance within the 10B parameter class on MMMU-Pro (40.7), MathVerse (42.6), and DynaMath (55.7). These results highlight the effectiveness of our dataset in enhancing the reasoning capabilities of vision-language models for complex multimodal tasks.
\end{abstract}    
\section{Introduction}
\label{sec:intro}
Vision-Language Models (VLMs), such as Llava~\citep{llava} and Gemini~\citep{gemini_1_5}, are designed to process multimodal inputs, including images, videos, and text. While VLMs have recently demonstrated significant progress in straightforward perceptual tasks such as VQA~\cite{antol2015vqa}, DocVQA~\cite{mathew2021docvqa}, and VizWiz~\citep{vizwiz}, they often struggle with more complex tasks such as MMMU~\cite{yue2024mmmu}, MathVista~\cite{lu2023mathvista}, and MEGA-Bench~\cite{chen2024mega}, which require multi-step, deliberate reasoning~\cite{wei2022chain, feng2023towards}. One major bottleneck for existing VLMs is the scarcity of reasoning-focused training datasets. Current multimodal reasoning datasets exhibit several limitations: (1) Many datasets, such as FigureQA~\cite{figureqa}, MapQA~\cite{chang2022mapqa}, GeoQA~\citep{GeoQA}, and ChartQA~\citep{masry2022chartqa}, focus narrowly on specific types of scientific images. (2) Some datasets rely on synthetic images generated through predefined rules, such as CLEVR~\citep{johnson2017clevr} and Geo170K~\citep{gllava}, which often result in poor generalization to real-world visual reasoning tasks. (3) Other training datasets, such as AI2D~\citep{ai2d} and ScienceQA~\citep{saikh2022scienceqa}, are relatively small and simplistic, primarily covering elementary-level visual knowledge. Due to these limitations, VLMs fail to acquire diverse reasoning skills, leading to slower progress on reasoning-intensive benchmarks compared to language models.
\begin{figure}
    \centering
    \includegraphics[width=1.05\linewidth]{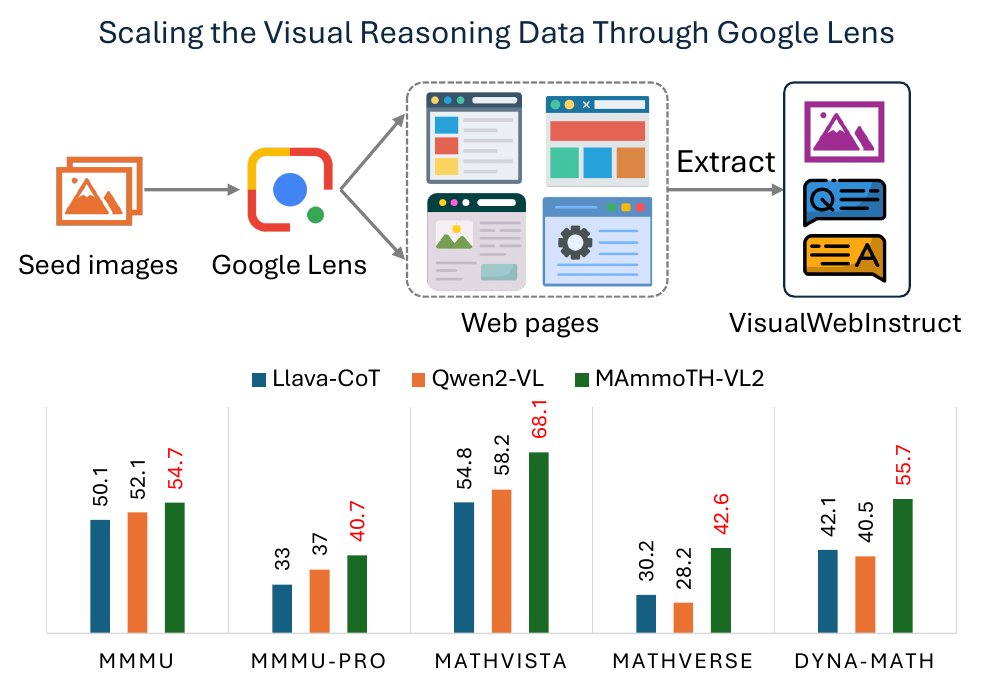}
    \caption{Overview of our automated data curation approach and major experimental results.}
    \label{fig:teaser}
\end{figure}

\begin{table*}[!t]
\small
\centering
\resizebox{\textwidth}{!}{
    \begin{tabular}{llll}
    \toprule
    Dataset   & Size & Source \& Domains                                 & Coverage                      \\
    \midrule
    ScienceQA~\cite{saikh2022scienceqa} & 21K & Elementary and high school science & Science Q\&A, diagrams, K-12 Exam \\
    IconQA~\cite{lu2021iconqa} & 107K & Abstract diagrams and visual reasoning & Visual reasoning, diagrams \\
    Geo170K~\citep{gllava}    & 170K & Synthesized from LLMs  & Geometry \\
    CLEVR~\citep{johnson2017clevr} & 700K & Synthesized from rules & Shapes  \\
    FigureQA~\citep{figureqa}  & 1.3M &  Synthesized from rules &  Bar, Line, Pie   \\
    ChartQA~\cite{masry2022chartqa} & 23K &  Charts from Staista, Pew, etc & Charts \\ 
    \midrule
    Math360V~\cite{shi2024mathllavabootstrappingmathematicalreasoning}  & 260K & FigureQA~\citep{figureqa}, CLEVR~\cite{johnson2017clevr}, IconQA~\citep{lu2021iconqa}, etc & Math reasoning, diagrams \\
    Mulberry~\cite{Mulberry} & 260K & Geo3K~\cite{lu2021inter}, IconQA~\cite{lu2021iconqa}, ChartQA~\cite{masry2022chartqa}, ScienceQA~\cite{saikh2022scienceqa}, etc & Geo, Figure, Medical, K-12 Exam \\
    Llava-CoT~\cite{llavacot} & 100K & ChartQA~\cite{masry2022chartqa}, AI2D~\cite{ai2d}, GeoQA~\cite{GeoQA}, CLEVR~\cite{johnson2017clevr}, etc & Geo, General VQA, K-12 Exam \\
    \midrule
    \data & 906K & Internet (Homework Website, Forums, etc) & All Above + College Exams \\
    \bottomrule
    \end{tabular}
}
\caption{Comparison between our dataset and the existing datasets. \data is the most diverse dataset with very broad coverage of disciplines and image types. }
\label{tab:dataset}
\end{table*}

Given the difficulty of human annotation, we draw inspiration from WebInstruct~\citep{mammoth2} to mine naturally existing reasoning-focused instruction data from the internet. While WebInstruct retrieves reasoning-focused text data from Common Crawl\footnote{https://commoncrawl.org/}, their approach is infeasible for the multimodal domain due to two key challenges: (1) the lack of a comparable large-scale multimodal dataset, and (2) the unreliability of current multimodal retrieval models. To address these challenges, we leverage commercial web image search tools, such as Google Image Search, which offer high coverage and accuracy.

We begin by collecting approximately 30,000 seed images across multiple disciplines, including Accounting, Chemistry, Mathematics, and Physics. These images serve as queries for Google Image Search~\cite{zhang2013image} to identify websites containing similar images. We then download the HTMLs from these websites and extract their accessibility trees, which are processed by an LLM to extract QA pairs (if any) for an initial dataset. However, we found that over half of the extracted questions lack annotated answers due to three primary reasons: (1) these websites do not provide answers, (2) some require membership to access, and (3) some necessitate user interaction to reveal the answers. To address this, we use GPT-4o~\citep{gpt4o} to synthesize multiple candidate solutions for each question, filtering for consistency among responses. Finally, we align the selected answers with the content from original webpage to remove potential inaccurate ones. Through this sophisticated process, we construct \data, a dataset containing approximately 900K QA pairs, where 40\% are visual QA pairs associated with 163,743 unique images, while the remaining 60\% are text-only QA pairs. Most of them are exam-like problems requiring deliberate reasoning.

Table~\ref{tab:dataset} compares \data with other datasets in terms of source and coverage. Our dataset comprises highly diverse, human-created scientific images spanning multiple disciplines and levels of complexity. Its broad coverage and increased difficulty make it particularly well-suited for improving VLM performance on real-world tasks requiring multi-step reasoning. To evaluate the effectiveness of \data, we perform supervised fine-tuning on MAmmoTH-VL~\citep{mammothvl} and Llava-OV-mid~\citep{llavaov}. Comprehensive evaluations across seven visual reasoning benchmarks, including MMMU~\citep{yue2024mmmu}, MathVista~\citep{lu2023mathvista}, and Dyna-Math~\citep{dynamath}, demonstrate substantial performance gains. When fine-tuning Llava-OV-mid, we observe an absolute improvement of 10--20 percentage points across these benchmarks. When fine-tuning MAmmoTH-VL, our model \model achieves state-of-the-art performance (within the 10B parameter range) on several benchmarks, including MMMU-Pro-std (40.7\%), MMVet (64.5\%), MathVerse (42.6\%), and Dyna-Math (55.7\%). \model's average performance across seven benchmarks surpasses strong competitors such as InternVL2.5~\citep{internvl_2_5} and Phi-4-Mini~\citep{phi_4_mini}, underscoring the effectiveness of \data in enhancing VLMs' reasoning capabilities.

\noindent Our contributions can be summarized as follows:
\begin{itemize}
\item We propose a scalable pipeline for acquiring high-quality multimodal reasoning data from the internet, ensuring both scalability and quality.
\item We introduce \data, a diverse and comprehensive multimodal instruction dataset, which we will publicly release to the research community.
\item We develop \model, a 7B-parameter vision-language model fine-tuned on \data, achieving state-of-the-art performance among models of comparable size and excelling in complex reasoning tasks requiring multi-step deliberation with visual context.
\end{itemize}

\noindent In the following sections, we will first talk about how we mine the data from the Internet in~\autoref{sec:stage1} and then talk about how to refine it in~\autoref{sec:stage2}. Finally, we show our experimental results in~\autoref{sec:exp}.
\begin{figure*}[!t]
    \centering
    \includegraphics[width=0.96\textwidth]{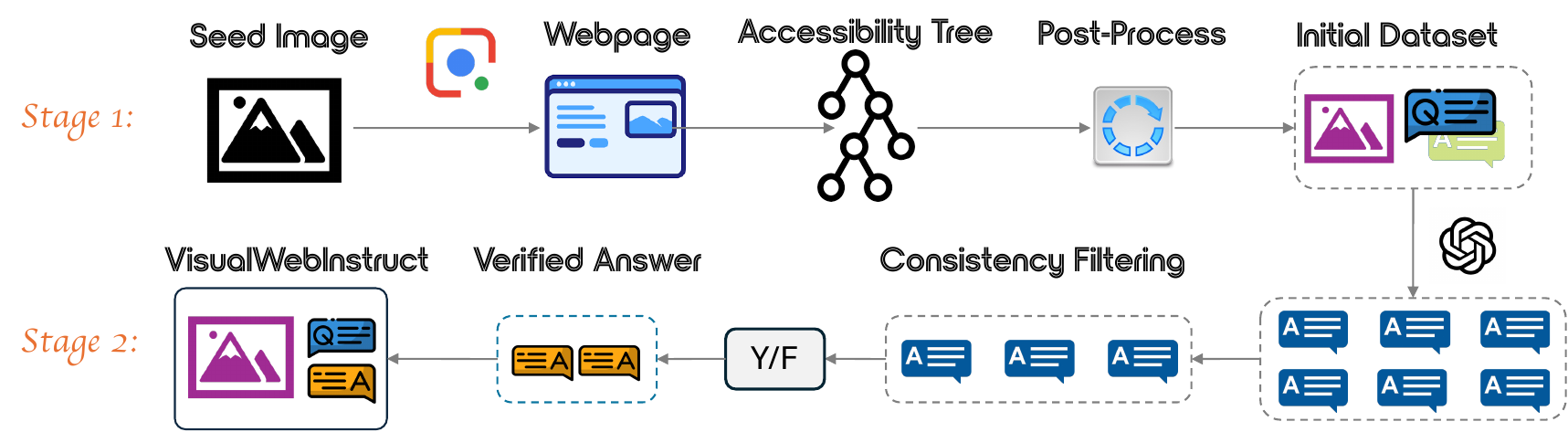}
    \caption{Comprehensive Pipeline for \data Dataset Generation. The workflow illustrates our multi-stage approach for creating high-quality multimodal instruction data. \textbf{Stage 1}: starting with seed images, we leverage Google Image search to identify relevant webpages, which are processed into accessibility trees. The raw QA pairs are extracted from the trees and refined through a post-processing step to ensure the vadality the data. \textbf{Stage 2}: we first generat multiple synthesized answers for consistency filtering, then align these with original web-sourced content to enhance the accuracy of the answers.}
    \label{fig:pipeline}
\end{figure*}

\section{Stage 1: Mining Data from the Internet}
\label{sec:stage1}
Our data mining pipeline follows a systematic approach to extract image-rich QA pairs from the internet. We begin with approximately 30K scientific images as seed data spanning multiple disciplines. We employ Google Image Search to identify visually similar content, gathering 758,490 unique URLs. After filtering out irrelevant domains, we construct accessibility trees for the relevant websites to extract meaningful content, preserving both textual and visual information while eliminating non-essential elements. We then leverage the Gemini 1.5 Flash model in a two-stage process: first to automatically extract QA pairs from the accessibility trees and then to filter these pairs based on comprehensive quality criteria, including question validity and image relevance, ensuring the educational value and integrity of the final dataset.

\subsection{Seed Data collecting}
Due to the limited availability of image-rich QA datasets and the predominant focus on mathematics in existing datasets, creating a comprehensive QA dataset that incorporates diverse subjects and abundant visual content is essential. Our seed dataset consists of approximately 30,000 images, which were crawled from Stemez\footnote{https://stemez.com/subjects/science/} in compliance with copyright regulations. These images span multiple disciplines, including mathematics, physics, accounting, chemistry, engineering, and biology, ensuring both subject diversity and visual richness.

\subsection{Google Image Searching}
\begin{figure}[htbp]
    \centering
    \includegraphics[width=0.5\textwidth]{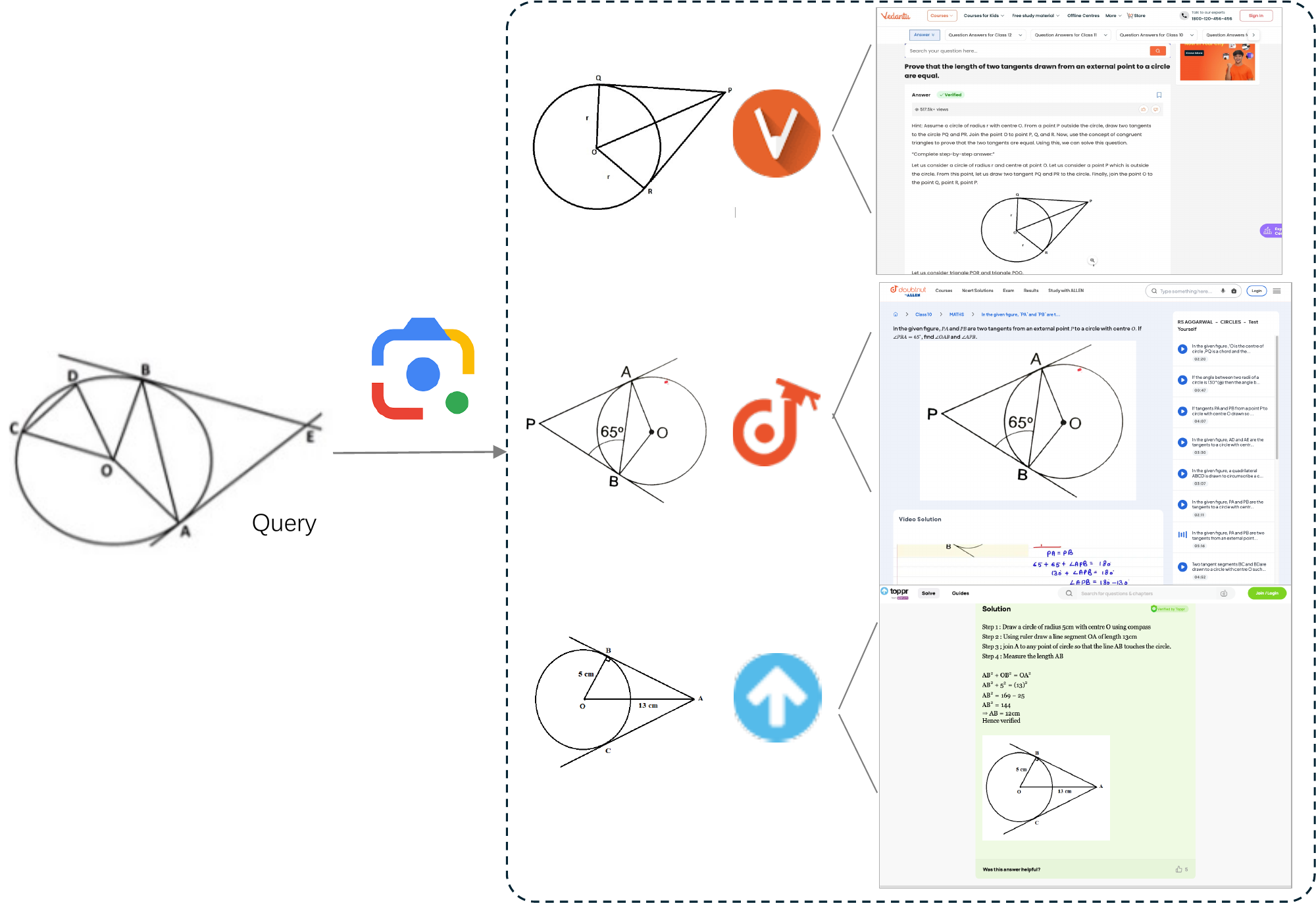}
    \caption{Example of Google Lens search functionality for circle geometry problems.}
    \label{fig:Image-Search}
\end{figure}

Using the seed images, we conducted Google Image searches to find visually similar content across the web. Leveraging Google Lens (Figure~\ref{fig:Image-Search}), we collected approximately 60 URLs per image, resulting in a total of 1,747,634 URLs containing visually similar content. Many websites with non-permissive licenses implement anti-crawling mechanisms, and we ensured compliance by avoiding data collection from such sources. We applied rigorous deduplication and filtering, removing URLs from domains unlikely to contain educational content (e.g., video platforms and image repositories). This refinement yielded 758,490 unique, high-quality URLs for further processing. By using images as primary search keys, we ensured strong visual and contextual connections between the collected data and our seed dataset, effectively preserving the original distribution while significantly expanding its coverage.

\subsection{Accessibility Tree Building}
\begin{figure}[htbp]
    \centering
    \includegraphics[width=0.4\textwidth]{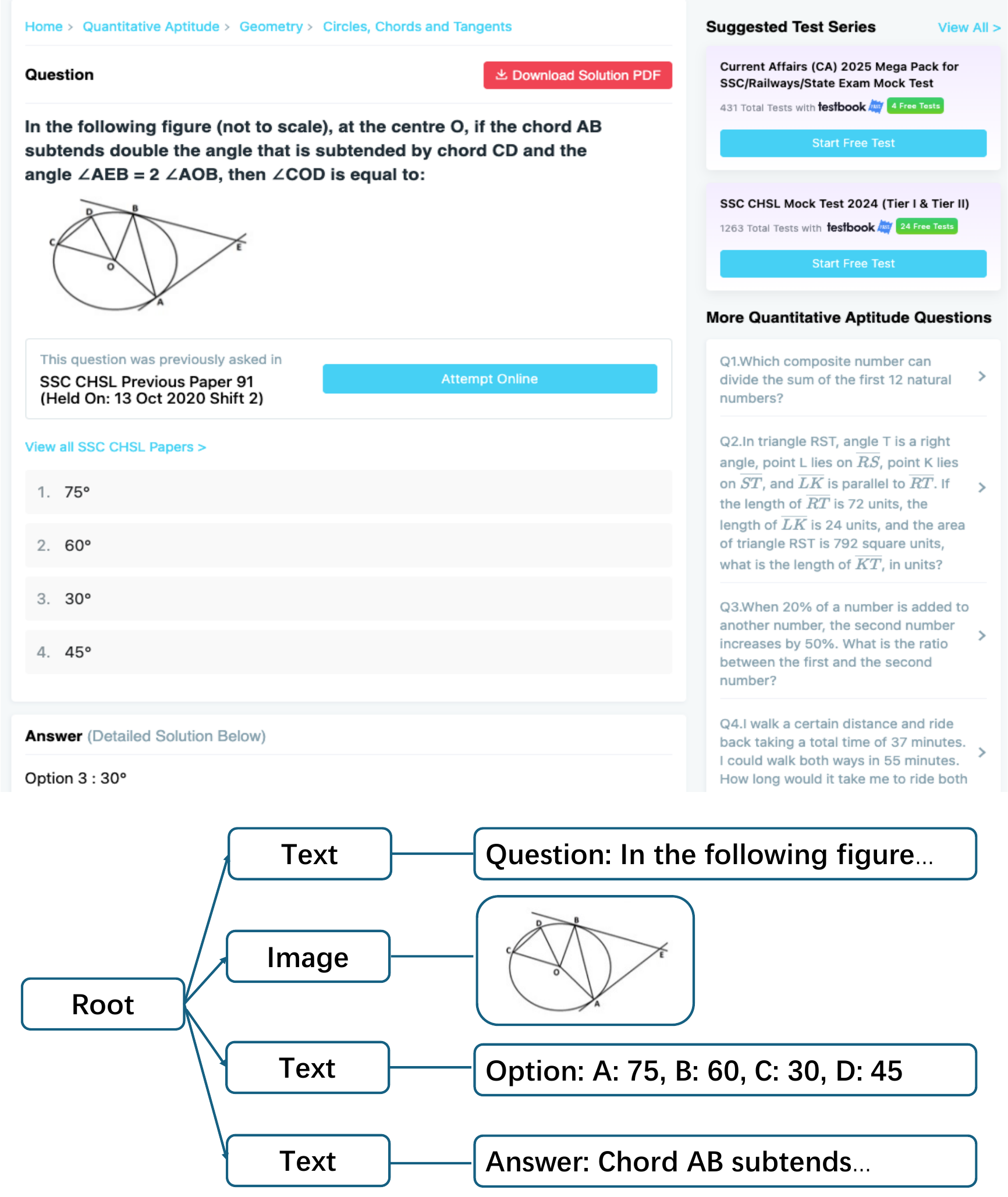}
    \caption{Example of an accessibility tree structure extracted from an educational website.}
    \label{fig:acc tree}
\end{figure}

After filtering out irrelevant domains, we processed the HTML content of each remaining URL to construct accessibility trees that capture essential textual and visual information. As illustrated in Figure~\ref{fig:acc tree}, our implementation focuses on extracting meaningful text content and image elements while filtering out non-essential components such as navigation menus, advertisements, and auxiliary elements. We developed a tree-based structure where each node represents either textual content or an image, preserving the hierarchical relationships present in the original HTML while removing unnecessary markup and styling information. The resulting accessibility trees provide a clean, hierarchical representation of each webpage's content, making subsequent QA pair extraction more efficient and reliable.

\subsection{QA Pairs Extraction}
After constructing accessibility trees, we prompt the Gemini 1.5 Flash model to identify and extract high-quality QA pairs from webpage content. We designed a structured prompt instructing the model to extract complete question text, identify relevant question-related images, and extract comprehensive solution details while preserving mathematical notations and step-by-step explanations. This approach maintains the educational integrity of the extracted content by preserving its original formatting, mathematical expressions, and logical structure, ensuring technical accuracy throughout the extraction process. Through this method, we extracted a total of 421,320 raw QA pairs from the webpages, with approximately 60\% containing images.

We then implemented a post-processing stage using the Gemini 1.5 Flash model to ensure dataset quality by evaluating both textual content and images. Our evaluation framework assessed two key criteria: question validity and meaningfulness, as well as the relevance and clarity of question-related images. By prompting Gemini to verify whether images are properly referenced, clear, visible, and contribute to understanding the question, we established strict validation criteria for retaining QA pairs.

This post-processing step significantly improved dataset quality by removing incomplete, unclear, or irrelevant content while preserving educational integrity and effectiveness. Our analysis shows that out of 421,320 processed pairs, 361,015 (85.7\%) were valid, while 60,305 were filtered out as invalid. Similarly, out of 449,859 total images processed, 331,818 (73.76\%) were deemed valid and relevant to their corresponding questions.
\section{Stage 2: Dataset Refinement}
\label{sec:stage2}

After Stage 1, we obtain a large amount of raw data from the Internet. However, this data contains a notable level of noise. For instance, more than half of the questions lack corresponding answers due to various issues, such as (1) membership requirements, (2) interaction requirements, and (3) the absence of an answer. Thus, a second round of refinement is necessary to further improve the dataset quality.

\subsection{Answer Refinement}
We implemented a comprehensive refinement process to ensure consistency and quality in our dataset. This step was critical in addressing potential variations or inconsistencies in the extracted answers, thereby creating a high-fidelity dataset for model training.

\begin{figure}[htbp]
    \centering
    \includegraphics[width=0.5\textwidth]{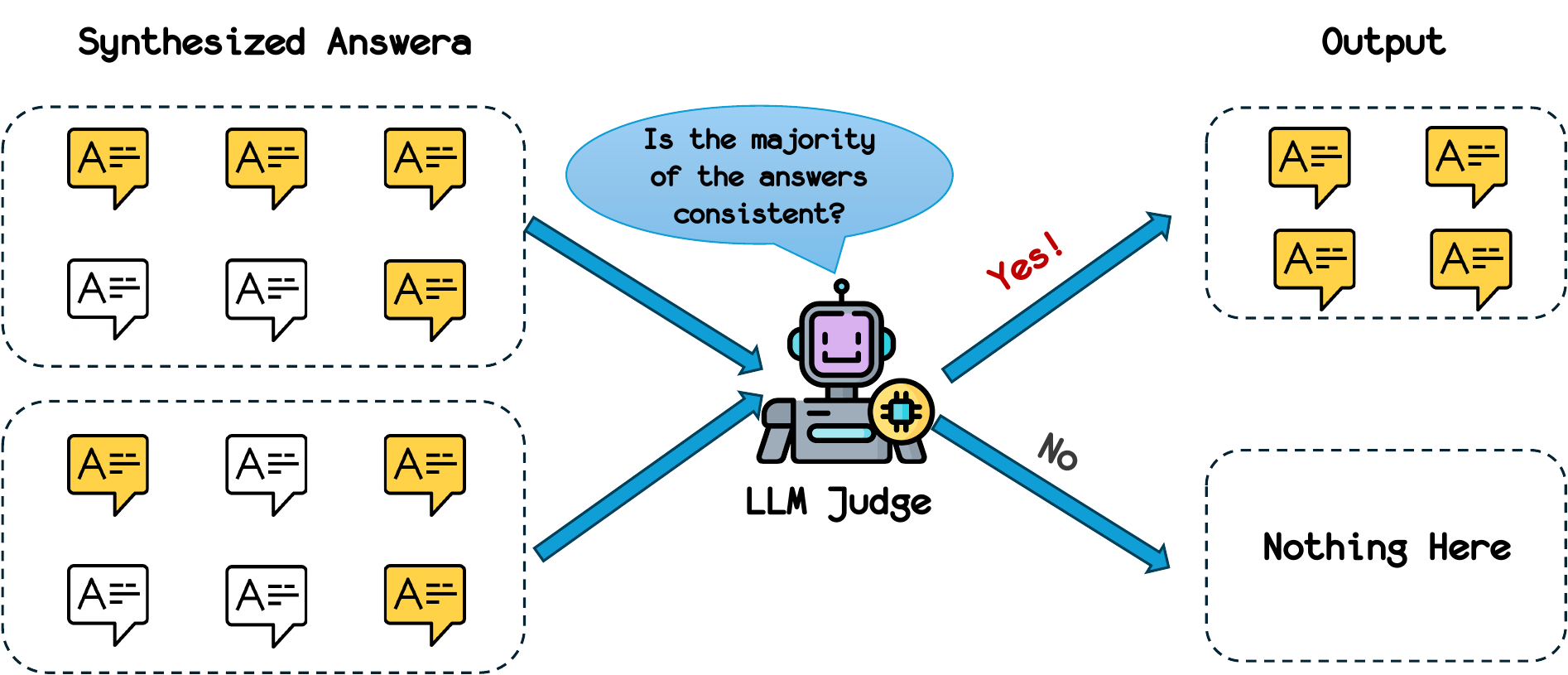}
    \caption{Illustration of our consistency checking methodology.}
    \label{fig:consistency checking}
\end{figure}

Our refinement methodology leveraged GPT-4o's capabilities in a two-stage process. First, for each question and its associated images, we prompted GPT-4o~\citep{gpt4o}\footnote{We compared GPT-4o and Gemini-1.5 and found that GPT-4o's outputs were significantly more reliable. Therefore, we adopted GPT-4o.} to generate four different answer variations. This approach allowed us to obtain multiple perspectives on each question. Next, we employed GPT-4o as an LLM judge to determine whether the synthesized responses aligned with each other. As illustrated in Figure~\ref{fig:consistency checking}, we evaluated whether the conclusions were mutually consistent across these responses. This evaluation was particularly important for questions in domains such as mathematics and physics, where precision and correctness are paramount. Only when more than half of the synthesized responses demonstrated consistency did we retain the question along with the consistent responses. This rigorous consistency check served as an additional quality filter, ensuring that our dataset contained highly accurate and unambiguous answers that could be reliably used for model training.

Through this refinement process, we successfully created a dataset in which all responses were systematically generated by GPT-4o, ensuring a consistent style and level of quality throughout the collection. The resulting dataset comprises 1.04 million QA pairs spanning multiple disciplines, representing one of the largest collections of consistency-verified multimodal instruction data available.

\subsection{Answer Alignment}
The final step in our quality assurance process involved answer alignment to further enhance accuracy. While the previous refinement step generated consistent answers using GPT-4o, we recognized the importance of validating these against authoritative content from the original web sources.

In this step, we used Gemini-2.0-Flash to measure the alignment between GPT-generated responses and the original extracted answers, if available. In cases where the comparison indicated inconsistency, we preserved the original web-sourced answer. Conversely, when the Gemini model determined strong alignment between the generated and web-sourced answers, we retained the GPT-generated version. Through this alignment process, we combined the consistency of model-generated content with the authority of original educational materials in a balanced manner.

\subsection{Dataset Statistics}
The statistics presented in Table~\ref{tab:category_distribution} illustrate the distribution of knowledge domains in our dataset, \data. While the major categories are shown in the table, the "Others" category (6.60\%) comprises General Knowledge (2.45\%), Computer Science (2.25\%), Biology (1.40\%), and humanities subjects, including Language/Literature (0.25\%), Social Sciences (0.20\%), and Arts (0.05\%). This distribution reflects the dataset's strong quantitative orientation while ensuring sufficient breadth. Table~\ref{tab:stage_stats} summarizes the statistics after each step of the \data pipeline, showing the data progression through two main stages. Our approach effectively scaled the initial 30,000 seed images into a comprehensive multimodal instruction dataset containing 900K instruction data. The final dataset includes 347,313 image-associated QA pairs (approximately 38\% of the total) supported by 163,743 unique images. 

We also conducted thorough decontamination checking to ensure our training dataset does not contain any data from the evaluation benchmarks, thereby maintaining the integrity of our experimental results.
\begin{table}[htbp]
\small
\centering
\begin{tabular}{lrr}
\toprule
\textbf{Category} & \textbf{Percentage}  & \textbf{Num of QA Pairs}\\
\midrule
Math & 62.50\%  & 566K\\
Physics & 14.50\% & 132K\\
Finance & 7.25\% & 66K\\
Chemistry & 4.80\% & 43K\\
Engineering & 4.35\% & 39K\\
Others & 6.60\% & 60K\\
\bottomrule
\end{tabular}
\caption{Distribution of Categories in \data}
\label{tab:category_distribution}
\end{table}
\begin{table*}[!t]
\centering
\small
\begin{tabular}{l|rrrrr}
\toprule
Processing Stage & Total QA Pairs & Image-Associated QA & Unique Questions & Total Images & Unique Images \\
\midrule
\multicolumn{6}{c}{Stage 1: Mining Data from the Internet} \\
\midrule
Seed Data Collection & - & - & - & 30,000 & 30,000 \\
QA Pairs Extraction & 421,320 & 248,643 & 421,320 & 552,269 & 362,728 \\
Post-Processing & 361,015 & 159,059 & 361,015 & 331,818 & 212,530 \\
\midrule
\multicolumn{6}{c}{Stage 2: Dataset Refinement} \\
\midrule
Answer Refinement & 1,041,598 & 407,218 & 257,201 & 577,455 & 167,493 \\
Answer Alignment & 906,160 & 347,313 & 257,201 & 475,099 & 163,743 \\
\bottomrule
\end{tabular}

\caption{Statistics of different milestones in the data processing pipeline of \data.}
\label{tab:stage_stats}
\end{table*}
\section{Experiments}
\label{sec:exp}
We detail the training and evaluation details of our experiments in this section. 
\subsection{Experimental Setup}
For our experiments, we directly fine-tuned an existing MAmmoTH-VL checkpoint on our \data dataset. We refer to our resulting model as \model. The architecture consists of a language tower based on Qwen2.5-7B-Instruct~\citep{qwen2_5}, a vision tower using SigLip~\citep{siglip}, and a projector module connecting these components, following Llava-OneVision~\citep{llava,llavaov}.
\begin{table*}[!t]
\centering
\small
\resizebox{0.9\textwidth}{!}{
\begin{tabular}{ll|cccccccc}
\toprule
Model       & Size & MMMU & MMMU-Pro & MMMU-Pro & MathVista & MMVet & MathVerse & Dyna-Math & Avg \\
            &      & val  & standard & vision & testmini & test & testmini & test &\\
\midrule
\multicolumn{9}{c}{Closed-sourced Models} \\
\midrule
GPT-4o      & -    & 69.1 & 54.0 & 49.7 & 63.8 & 76.2 & 50.2 &  63.7 &  61.0 \\
Gemini-1.5-Pro & - & 59.1 & 49.4 & 65.8  & 63.9 & 64.0 & 41.2 & 64.8 &  58.3 \\
Claude-3.5-Sonnet & - & 68.3 & 55.0 & 48.0 & 67.7 & 75.4 & 44.2 & 60.5 & 59.9\\ 
\midrule
\multicolumn{9}{c}{Open-source General Vision-Language Models} \\
\midrule
Molmo       & 8B   & 45.3 & 28.3 & 18.9 & 51.6 & 58.0  & 18.9 & 41.6 & 37.5\\
Llava-OV    & 7B   & 48.8 & 29.5 & 18.7 & 63.2 & 58.6  & 26.2 & 40.3 & 40.8\\      
Llama-3.2-Inst  & 11B  & 50.7 & 33.0 & 23.7 & 51.5 & 59.3  & 31.6 &  40.5 & 41.5\\
Qwen2-VL    & 7B   & 52.1 & 37.0 & 26.9 & 58.2 & 62.0  & 28.2 & 42.1 & 43.8\\
MAmmoTH-VL  & 7B   & 50.8 & 33.2 & 25.3 & \secondbest{66.0} & 62.3  & 34.2 &  44.7 & 45.2\\
InternVL2.5 & 7B   & \best{55.8} & 38.2 & \secondbest{30.4} & 64.4 & 62.8  & \secondbest{39.5} & 49.8 & \secondbest{48.7}\\
Phi-4-mini  & 5.6B & \secondbest{55.1} & \secondbest{39.7} & \best{31.2} & 62.4 & 60.5  & 37.6 & \secondbest{51.4} & 48.6\\
DeepSeek-VL2  & 27B & 51.1 & 31.4 & 24.3 & 62.8  & - & -  & - & -\\ 
Llava-CoT-L  & 11B  & 50.1 & 31.6 & 20.4 & 54.8 & 60.3  & 30.2 & 44.8 & 41.7  \\
Llava-CoT-M   & 7B   & 51.4 & 33.0 & 23.7 & 63.8 & 58.6  & 39.4 & 48.3 & 45.5 \\
LlamaV-o1    & 11B  & 49.1 & 31.5 & 22.4 & 54.4 & \secondbest{63.6}  & -    &  -   & -    \\
Mulberry     & 7B   & 55.0 & 36.8 & 23.6 & 63.1 & 60.9  & 31.0 & 45.1 & 45.0 \\
Insight-V    & 8B   & 50.2 & 30.7 & 20.5 & 59.9 & 60.8  & 28.7 & 47.8 & 42.6 \\
MM-Eureka    & 8B   & 49.2 & -    & -    & 67.1 & 60.7  & 40.4 & -    & -     \\ 
\midrule
\model         & 7B   & 54.7 & \best{40.7} & 26.3 & \best{68.1} & \best{64.5}  & \best{42.6} & \best{55.7} & \best{50.4} \\
$\Delta$ over SoTA  &      & -1.1 & +1.0 & -4.9 & +2.1 & +0.9  & +3.1 & +4.3 & +1.7 \\
\bottomrule
\end{tabular}
}
\caption{Evaluation Results of our model and other baseline models. Most of the baseline results are taken from other papers. The \textbf{best} and \underline{second-best} results across all open-source models are highlighted in bold and underlined, respectively.}
\label{tab:Model performance}
\end{table*}

To enhance data diversity, we employed a data mixing strategy that combined our \data dataset with modified LLaVA-CoT~\citep{llavacot} (with CoT prompting tags removed) in a 9:1 ratio, resulting in approximately 900K samples from \data and 100K samples from the modified LLaVA-CoT dataset. This mixing strategy empirically improved our model's performance across diverse visual reasoning tasks.

We employed a supervised fine-tuning (SFT) approach with a batch size of 256. The learning rate was set to $1 \times 10^{-5}$ for the language model and projector components, while the vision encoder was fine-tuned with a lower rate of $2 \times 10^{-6}$ to preserve its pre-trained visual recognition capabilities. The model was trained for a single epoch, which proved sufficient given the high quality and diversity of our dataset. Input images were processed at a resolution of $384 \times 384$ with appropriate adjustments for varied aspect ratios. We limited input sequences to a maximum of 8,192 tokens to accommodate detailed reasoning chains while maintaining computational efficiency.

This fine-tuning approach enabled \model to leverage the strong multimodal reasoning foundation of MAmmoTH-VL while enhancing its performance on our targeted visual reasoning tasks that require multi-step deliberation with visual context.
\subsection{Evaluation Setup}
\label{subsec:evaluation_setup}

To assess the capabilities of \model, we conducted a comprehensive evaluation across multiple multimodal benchmarks that specifically test visual reasoning and knowledge application. Our evaluation framework focuses on benchmarks that require complex reasoning with visual context. We evaluate our model on seven key benchmarks that collectively provide a comprehensive assessment of multimodal reasoning capabilities:

\begin{itemize}
    \item \textbf{MMMU~\citep{yue2024mmmu}}: Tests multimodal understanding across university-level domains, requiring integration of visual and textual information.
    
    \item \textbf{MMMU-Pro~\citep{mmmupro}}: Advanced versions of MMMU with more challenging problems and more distractor options that require sophisticated visual reasoning.
    
    \item \textbf{MathVista~\citep{lu2023mathvista}}: Evaluates mathematical reasoning with visual inputs, testing the model's ability to process visual information for solving complex math problems.
    
    \item \textbf{MMVet~\citep{mmvet}}: Assesses general multimodal understanding across diverse tasks and contexts.
    
    \item \textbf{MathVerse~\citep{mathverse}}: Focuses on mathematical reasoning with visual components and relies less on text hints, requiring complex visual reasoning.
    
    \item \textbf{Dynamath~\citep{dynamath}}: Tests dynamic mathematical reasoning capabilities with visual context.
\end{itemize}
For all evaluations, we used greedy decoding in a zero-shot setting to ensure fair comparison with existing models. We categorize the comparison models into three groups: 
\begin{itemize}
    \item Closed-source Models: GPT-4o~\citep{gpt4o}, Gemini-1.5-Pro~\citep{gemini_1_5}, Claude-3.5-Sonnet~\citep{claude35}.
    \item Open-source Vision-Language Models: Molmo~\citep{molmo}, LLaVA-OV~\citep{llavaov}, Llama-3.2~\citep{llama3}, Qwen2-VL~\citep{qwen2vl}, MAmmoTH-VL~\citep{mammothvl}, InternVL2.5~\citep{internvl_2_5}, Phi-4-mini~\citep{phi_4_mini}, DeepSeek-VL2~\citep{deepseekvl2}.
    \item Reasoning Vision-Language Models: SFT models like Llava-CoT-L (from Llama-3.2) and Llava-CoT-M (from MAmmoTH-VL)~\citep{llavacot}, LLama-V-o1~\citep{llamav_o1}, Mulberry~\citep{Mulberry}, Insight-V~\citep{insight_v}. We include a recent work MM-Eureka~\citep{mmeureka} trained with RL. 
\end{itemize}
To ensure standardized and reproducible evaluations, we employed LMMsEval~\cite{zhang2024lmmsevalrealitycheckevaluation}, a comprehensive evaluation framework for multimodal language models. For all evaluations, we used greedy decoding in a zero-shot setting to ensure fair comparison with existing models. Our approach allows for direct comparison with models of comparable size, providing insights into the value of the \data dataset. Performance is reported using accuracy scores for each benchmark, with an average score across all benchmarks to indicate overall model capability.

\begin{table*}[!t]
\centering
\small
\begin{tabular}{l|cccccccc}
\toprule
Training Data & MMMU & MMMU-Pro & MMMU-Pro & MathVista & MMVet & MathVerse & Dyna-Math & Avg\\
& val  & standard & vision & testmini & test & testmini & test &\\
\midrule
\multicolumn{9}{c}{Training from LLava-OV-mid} \\
\midrule
-        & 40.1 & 21.2 & 12.2 & 36.0 & 32.1 & 18.1 & 24.4 & 26.3 \\
Llava-CoT & 40.8 & 25.8 & 14.6 & 45.7 & 47.5 & 27.2 & 33.9 & 33.6 \\
Ours & 45.3 & 31.5 & \best{20.9} & 43.9 & \best{57.6} & 27.4 & 40.3 & 38.1 \\
Ours+Llava-CoT  & \best{47.6} & \best{31.6} & \best{20.9} & \best{48.8} & 51.7 & \best{34.9} & \best{42.3} & \best{39.7} \\
\midrule
\multicolumn{9}{c}{Training from MAmmoTH-VL} \\
\midrule
-        & 50.8 & 34.8 & 25.3 & 66.0 & 62.3 & 34.2 & 44.7 & 45.4 \\
Llava-CoT & 51.4 & 35.2 & 24.6 & 63.8 & 58.7 & 39.4 & 48.3 & 45.9 \\
Ours & 52.6 & 38.6 & \best{29.0} & 65.9 & 61.8 & 39.4 & 55.7 & 49.0 \\
Ours+Llava-CoT  & \best{54.7} & \best{40.7} & 26.3 & \best{68.1} & \best{64.5} & \best{42.6} & \best{55.7} & \best{50.4} \\
\bottomrule
\end{tabular}
\caption{Ablation Results of our experiments. We show experimental results from different backbones to show the impact of consistency filtering and data mixing with Llava-CoT. For each base model, the \best{best} performance is highlighted in bold.}
\label{tab:Model performance ablation}
\end{table*}

\subsection{Experimental Results}
Here we evaluate our results from different perspectives.

\noindent \textbf{Quantitative Results}
The table \ref{tab:Model performance} presents the performance of \model compared to various multimodal models across seven benchmarks. Our analysis reveals several important findings regarding the effectiveness of models fine-tuned on \data.

\noindent \textbf{Overall Performance.} \model achieves an average accuracy of 50.4\% across all benchmarks, outperforming other open-source vision-language models of comparable size (7B-11B parameters). This represents a significant improvement over standard vision-language models like Qwen2-VL (43.8\%), LLaVA-OV (40.8\%), and Molmo (37.5\%). It even beats the very recent model like InternVL2.5~\citep{internvl_2_5} and Phi-4-mini-Multimodal~\citep{phi_4_mini}.

\noindent \textbf{Mathematical Reasoning Capabilities.} \model demonstrates particularly strong performance on mathematical reasoning tasks. On MathVista, our model achieves 68.1\% accuracy, surpassing all the open-source and closed-source models. The model's performance on MathVerse (42.6\%) and Dyna-Math (55.7\%) further confirms its enhanced capability for visual reasoning.

\noindent \textbf{Complex Reasoning Tasks.} On MMMU-Pro-std with 10 options, \model achieves 40.7\% accuracy, showing a significant improvement over other 7B models such as LLaVA-OV (29.5\%) and Qwen2-VL (37.0\%). This demonstrates that our approach effectively enhances the model's ability to perform complex reasoning across diverse domains beyond mathematics.

\noindent \textbf{Gap with Larger and Closed-Source Models.} While \model outperforms open-source models of comparable size, there remains a gap with closed-source models such as GPT-4o, Gemini-1.5-Pro, and Claude-3.5-Sonnet. This indicates potential for further improvements through scaling or enhanced training methodologies.

\noindent \textbf{Comparison with Reasoning-Enhanced Models.} Among the reasoning-enhanced vision-language models like Llava-CoT, Mulberry~\citep{Mulberry}, LlamaV-o1~\citep{llamav_o1} and Insight-V~\citep{insight_v}, \model demonstrates competitive performance, achieving results comparable to or better than specialized models like LLaVA-CoT and Mulberry. For instance, on MMMU-Pro Vision, our model achieves 26.3\% accuracy, outperforming LLaVA-CoTM's 23.7\%. Notably, other reasoning-enhanced models often utilize complex methodologies in either the training or inference stage to enhance their chain-of-thought abilities, which makes the development process and deployment more complicated. In contrast, \model achieves much better reasoning capabilities through our straightforward fine-tuning approach on \data, offering a simpler yet effective solution compared to the other approaches.

These results confirm that fine-tuning on \data significantly enhances the model's reasoning capabilities. The consistent performance improvements across diverse benchmarks from non math-related and math-related domains demonstrate the effectiveness of our approach in developing more capable multimodal reasoning models. We believe our dataset can be utilized to augment future vision-language models.

\subsection{Ablation Study}
The ablation study in Table \ref{tab:Model performance ablation} demonstrates the impact of different training datasets and their combinations on model performance across multiple visual reasoning benchmarks. Two base models were evaluated: Llava-OV-mid and MAmmoTH-VL.

For Llava-OV-mid, the baseline starts at 26.3\% average score across benchmarks. Training with Llava-CoT data improves this to 33.6\%, while training on \data yields an even better 38.1\%, with with MMVet performance notably jumping from 32.1\% to 57.6\%. The combined training approach (\data+Llava-CoT) achieves the best overall performance at 39.7\%.

The stronger MAmmoTH-VL model begins with an average score of 45.4\%. Training with \data improves the average to 49.0\%, showing gains across multiple benchmarks, particularly in MMMU-Pro vision and Dyna-Math tests. As with Llava-OV-mid, the combined training approach works best, reaching 50.4\% average score, with notable improvements in MMMU (54.7\%), MMMU-Pro standard (40.7\%), and Dyna-Math (55.7\%).

The key findings indicate strong data complementarity between \data and Llava-CoT, with their combination consistently delivering the best results. We also observe that weaker base models show larger relative improvements from training. Overall, the ablation study confirms that our \data dataset significantly boosts model performance across all benchmarks, demonstrating its effectiveness in enhancing visual reasoning capabilities regardless of the base model. 
\section{Related Works}

\subsection{Multimodal Instruction Data}
Creating high-quality multimodal datasets remains a significant challenge in advancing MLLMs. Current approaches face critical limitations, particularly in balancing quality and scale. Human-annotated datasets provide high-precision, contextually appropriate data~\cite{xu2024visionflanscalinghumanlabeledtasks, molmo, mckinzie2024mm1methodsanalysis, sun2023aligninglargemultimodalmodels} but suffer from prohibitive costs and scalability constraints. Meanwhile, methods leveraging existing academic datasets~\cite{tong2024cambrian1fullyopenvisioncentric, liu2023gptunderstands} offer more cost-effective alternatives but lack the diversity and reasoning complexity needed for advanced multimodal reasoning tasks. This limitation is particularly evident in the scarcity of large-scale, reasoning-focused multimodal datasets that can be efficiently produced. Our work addresses these challenges by proposing a novel, scalable methodology for constructing multimodal instruction datasets that maintain both the quality and reasoning complexity.

\subsection{Multimodal Large Language Models}
Multimodal Large Language Models (MLLMs) have advanced AI by integrating text and visual processing capabilities. While proprietary models such as GPT-4o~\cite{gpt4o} and Gemini~\cite{gemini_1_5, team2024gemini} achieve state-of-the-art performance, they remain inaccessible to the broader research community. To address this gap, connector-based approaches~\cite{li2023blip2bootstrappinglanguageimagepretraining,dai2023instructblipgeneralpurposevisionlanguagemodels} have emerged, linking visual encoders to language models through lightweight projection modules.

Recent open-source MLLMs, such as LLAMA~\cite{llama3}, LLaVA~\cite{li2024llava,liu2023visualinstructiontuning}, MiniGPT-4~\cite{zhu2023minigpt4enhancingvisionlanguageunderstanding}, and Deepseek-VL~\cite{lu2024deepseekvlrealworldvisionlanguageunderstanding}, have contributed to advancements in vision-language understanding. Additionally, Qwen-VL~\cite{qwen2vl} and InternVL~\citep{internvl_2_5} have demonstrated strong performance through efficient design and diverse pre-training.

Meanwhile, various approaches have been developed to enhance MLLM reasoning capabilities, including neural symbolic methods~\cite{amizadeh2020neurosymbolicvisualreasoningdisentangling,Choi_2024}, optimized visual encoding strategies~\cite{jin2024chatuniviunifiedvisualrepresentation,li2024tokenpackerefficientvisualprojector}, plan-based prompting~\cite{mitra2024compositionalchainofthoughtpromptinglarge,luan2024textcotzoomenhancedmultimodal}, structured reasoning frameworks~\cite{llavacot}, and sequential instruction tuning~\cite{hu2024finetuninglargelanguagemodels}. Despite these advancements, these models face a critical challenge: the scarcity of publicly available large-scale visual reasoning datasets necessary for enhancing model reasoning capabilities~\cite{bai2024surveymultimodallargelanguage}. Our work addresses this supervised fine-tuning data bottleneck while building on the connector-training paradigm, aiming to bridge the gap between proprietary and open-source multimodal models to foster more accessible vision-language systems.

\subsection{Chain-of-Thought in Large Language Models}
Chain-of-Thought (CoT) prompting~\cite{wei2023chainofthoughtpromptingelicitsreasoning} has revolutionized how large language models tackle complex reasoning challenges. This technique enables LLMs to navigate difficult problems—including commonsense scenarios~\cite{geva2021didaristotleuselaptop,sap-etal-2020-commonsense} and logical puzzles~\cite{10141570,xiong2024teilptimepredictionknowledge}—by following explicit reasoning pathways. At its core, CoT methodically decomposes complex questions into manageable sequential steps, creating a structured framework that guides models toward systematic solutions~\cite{chu2024navigateenigmaticlabyrinthsurvey}. Evidence consistently demonstrates significant improvements in reasoning performance through this approach. Notable advancements include Prism~\cite{qiao2024prismframeworkdecouplingassessing}, which implements a distinctive dual-stage architecture that separates initial perception from subsequent reasoning operations, and MSG~\cite{cesista2025multimodalstructuredgenerationcvprs}, which pioneered the forced Chain-of-Thought methodology—establishing a foundational paradigm shift in structured prompting approaches that continues to shape current research.
\section{Conclusion}
In this paper, we explore the possibility of constructing large-scale multimodal reasoning datasets without relying on human annotation. We are the first paper to utilize Google Image Search for mining high-quality visual reasoning dataset. Our approach has been highly effective to achieve state-of-the-art performance on 5 out of 7 evaluated benchmarks. In the future, we plan to work on multiple round of search to further expand the dataset size.
\section*{Acknowledgement}
This research was supported by NetMind.Ai for providing cloud compute. Also, we also want to thank Google DeepMind for generous support for Gemini credits. A large part of our data processing pipeline is benefited from the credits.
{
    \small
    \bibliographystyle{ieeenat_fullname}
    \bibliography{main}
}

\clearpage
\section{Supplementary Material}

\subsection{Image Number Distribution}
\begin{figure}[htbp]
    \centering
    \includegraphics[width=0.6\textwidth]{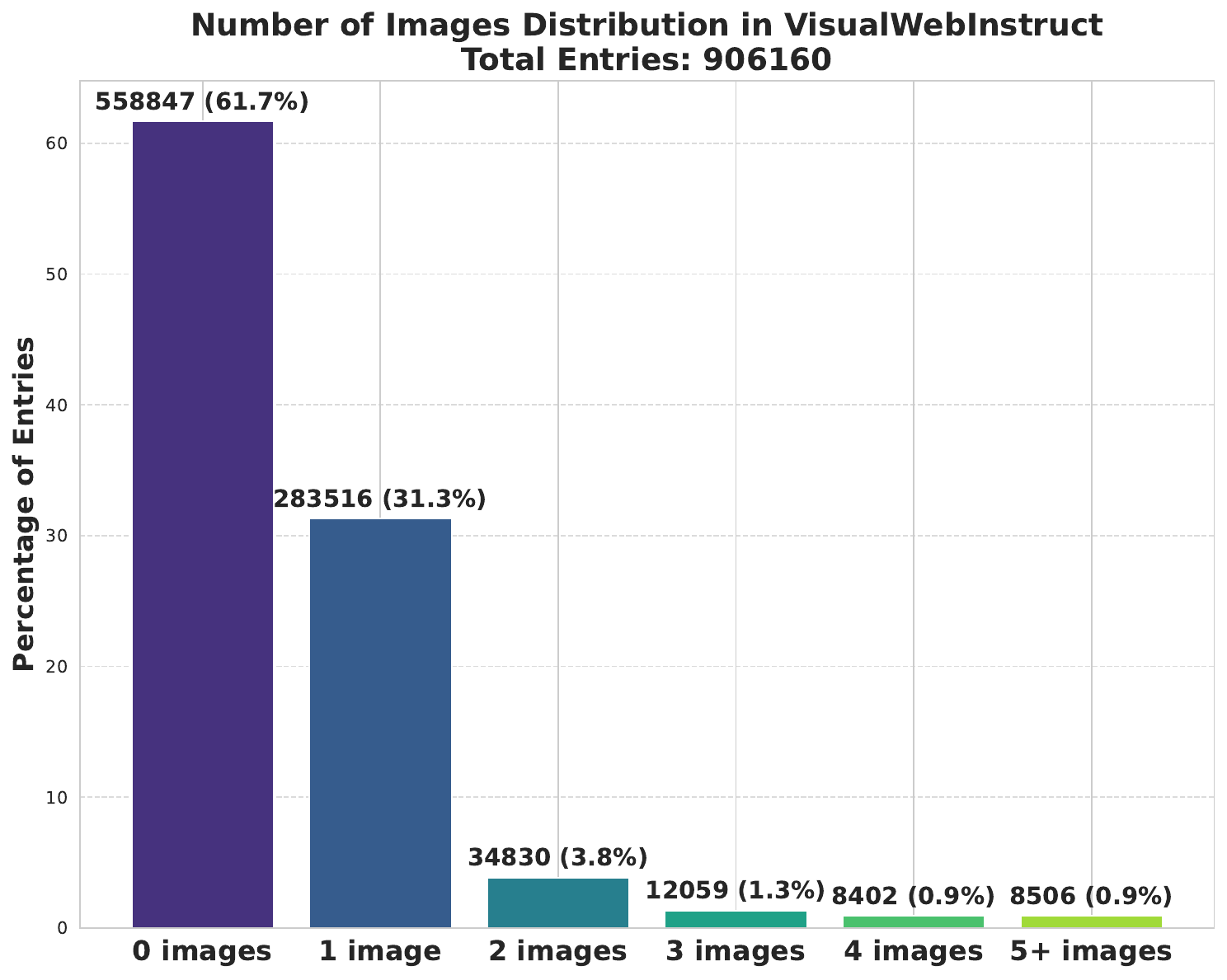}
\end{figure}
\subsection{\model Training Configuration}
\begin{table}[htbp]
\centering
\small
\begin{tabular}{l|l||l|l}
\hline
\multicolumn{2}{c||}{\textbf{Model Architecture}} & \multicolumn{2}{c}{\textbf{Data Processing}} \\
\hline
Base Language Model & Qwen/Qwen2.5-7B-Instruct & Image Aspect Ratio & anyres\_max\_4 \\
Vision Encoder & google/siglip-so400m-patch14-384 & Image Grid Pinpoints & (1x1),...,(6x6) \\
Vision-Language Connector & MLP-based projector (2-layer with GELU) & Group by Modality & Enabled \\
Vision Select Layer & -2 (second-to-last layer) & Image Start/End Tokens & Disabled \\
Patch Merge Type & spatial\_unpad & Image Patch Token & Disabled \\
Starting Checkpoint & MAmmoTH-VL & Lazy Preprocessing & Enabled \\
\hline
\multicolumn{2}{c||}{\textbf{Training Configuration}} & \multicolumn{2}{c}{\textbf{Dataset Configuration}} \\
\hline
Training Epochs & 1 & Primary Dataset & VisualWebInstruct \\
Batch Size & 256 & Additional Dataset & LLaVA-CoT (9:1 ratio) \\
Maximum Sequence Length & 8,192 tokens & Prompt Template & qwen\_2\_5 \\
Learning Rate & 1e-5 (language and projector) &  & \\
Vision Tower Learning Rate & 2e-6 &  & \\
Weight Decay & 0.0 &  & \\
Warmup Ratio & 0.03 &  & \\
LR Scheduler & Cosine &  & \\
\hline
\multicolumn{2}{c||}{\textbf{Tunable Components}} & \multicolumn{2}{c}{\textbf{Optimization}} \\
\hline
Language Model & Enabled & Distributed Training & DeepSpeed Zero-3 \\
Vision Tower & Enabled & TF32 Precision & Enabled \\
MLP Adapter & Enabled & Mixed Precision & BF16 \\
Gradient Checkpointing & Enabled & TF32 Precision & Enabled \\
Torch Compile & Enabled (inductor) &  &  \\
\hline
\end{tabular}
\label{tab:training_config}
\end{table}

\clearpage
\subsection{Prompt for Each Stage}
\begin{figure}[htbp]
\centering
\setlength{\fboxsep}{10pt}
\setlength{\fboxrule}{0.5pt}
\fbox{%
\begin{minipage}{0.95\textwidth}
\small
\textbf{QA Pairs Extraction}

\vspace{1ex}
\ttfamily
"""Analyze this webpage content and extract questions, images, and\\{} 
complete solution details in Markdown format.\\{}
Please format your response as follows:\\{}
**Question 1:**\\{}
[complete question text]
\\{}
**Images:**\\{}
* [First image URL if available]
\\{}
* [Second image URL if available]
\\{}
[continue for each additional image...]
\\{}
**Solution:**\\{}
[Copy the complete solution text from the webpage, including all steps,\\{}
explanations, and calculations]
\\{}
**Images in Solution:**\\{}
* [First image URL if available]
\\{}
* [Second image URL if available]
\\{}
[continue for each additional image...]
\\{}
[repeat for each additional question...]
\\{}
Requirements:\\{}
- Keep the complete solution text exactly as shown in the webpage\\{}
- Use Markdown formatting throughout the response\\{}
- Mark missing content as "Not found"\\{}
- For images, include URL only\\{}
- For multiple questions, number them sequentially\\{}
- Do not summarize or modify the solution text\\{}
- Preserve all mathematical notations and formulas\\{}
- Keep all step-by-step explanations intact\\{}
- Preserve all line breaks and indentation in solution text\\{}
- If there is no question in the content, mark it as "Not found"\\{}
- If the webpage is empty or missing, return nothing\\{}
Webpage content:\\{}
\{Accessibility Tree\}\\{}
"""
\end{minipage}%
}
\end{figure}

\begin{figure*}[htbp]
\centering
\setlength{\fboxsep}{10pt}
\setlength{\fboxrule}{0.5pt}
\fbox{%
\begin{minipage}{0.95\textwidth}
\small
\textbf{QA Pairs Validation}

\vspace{1ex}
\ttfamily
"""Please analyze this question-answer pair and its images:\\{}
Question: complete question text\\{}
Solution: complete solution text\\{}
Your tasks:\\{}
1. Determine if the question is meaningful and valid.\\{}
2. For the question images (if any), determine if each is:\\{}
   - Properly referenced in the question\\{}
   - Clear and visible\\{}
   - Actually helps understand the question\\{}

3. For the solution images (if any), determine if each is:\\{}
   - Helps explain the solution\\{}

Notes:\\{}
- Image indices start from 0 (e.g., first image is index 0, second is index 1, etc.)\\{}
- Images should be marked as valid if they show the actual content being discussed\\{}
- Images should be marked as invalid only if they are:\\{}
  * Completely irrelevant to the question/solution\\{}
  * Corrupted or unreadable\\{}
  * Duplicate or redundant\\{}

Question Images:\\{}
[Images loaded here]
Solution Images (starting a new section, indexes reset to 0):\\{}
[Images loaded here]
Please respond in this exact format:\\{}
QUESTION\_VALID: [yes/no]
\\{}
ANALYSIS: [Brief explanation of why the question is valid/invalid]
\\{}
QUESTION\_IMAGES: [comma-separated list of valid image indices starting from 0]
\\{}
QUESTION\_IMAGES\_REASON: [Brief explanation for each image decision]
\\{}
SOLUTION\_IMAGES: [comma-separated list of valid image indices starting from 0]
\\{}
SOLUTION\_IMAGES\_REASON: [Brief explanation for each image decision]
\\{}

CRITICAL RESPONSE FORMAT INSTRUCTIONS:\\{}
- You MUST respond using EXACTLY this format with no additional text\\{}
- Use ONLY numeric indices for images, starting from 0\\{}
- If no images are valid, use an empty string\\{}
- Be precise and use actual numbers\\{}
- Always use numeric indices (0,1,2...)\\{}
- Use empty string for no images (e.g., "SOLUTION\_IMAGES: ")\\{}
- Do not add explanatory text in the indices field\\{}
"""
\end{minipage}%
}
\end{figure*}

\begin{figure}[htbp]
\centering
\setlength{\fboxsep}{10pt}
\setlength{\fboxrule}{0.5pt}
\fbox{%
\begin{minipage}{0.95\textwidth}
\small
\textbf{Answer Alignment}

\vspace{1ex}
\ttfamily
"""Given the question and the provided image(s), compare these two answers and determine if they are aligned.

Question: {question}

GPT's Answer: {gptanswer}

Real Answer: {realanswer}

Example of Aligned Answers: \\{}
Question: What is 2 + 2? \\{}
GPT Answer: 4 \\{}
Real Answer: 4

Example of Misaligned Answers: \\{}
Question: What is derivative of $x^2$? \\{}
GPT Answer: 2x + 1 \\{}
Real Answer: 2x

Are these answers aligned? Respond with just 'Yes' or 'No' on the first line. Provide a brief explanation on the second line."""
\end{minipage}%
}
\end{figure}
\clearpage
\subsection{Example of QA Pair}
\begin{figure}[!htbp]
    \centering
    \includegraphics[width=\textwidth]{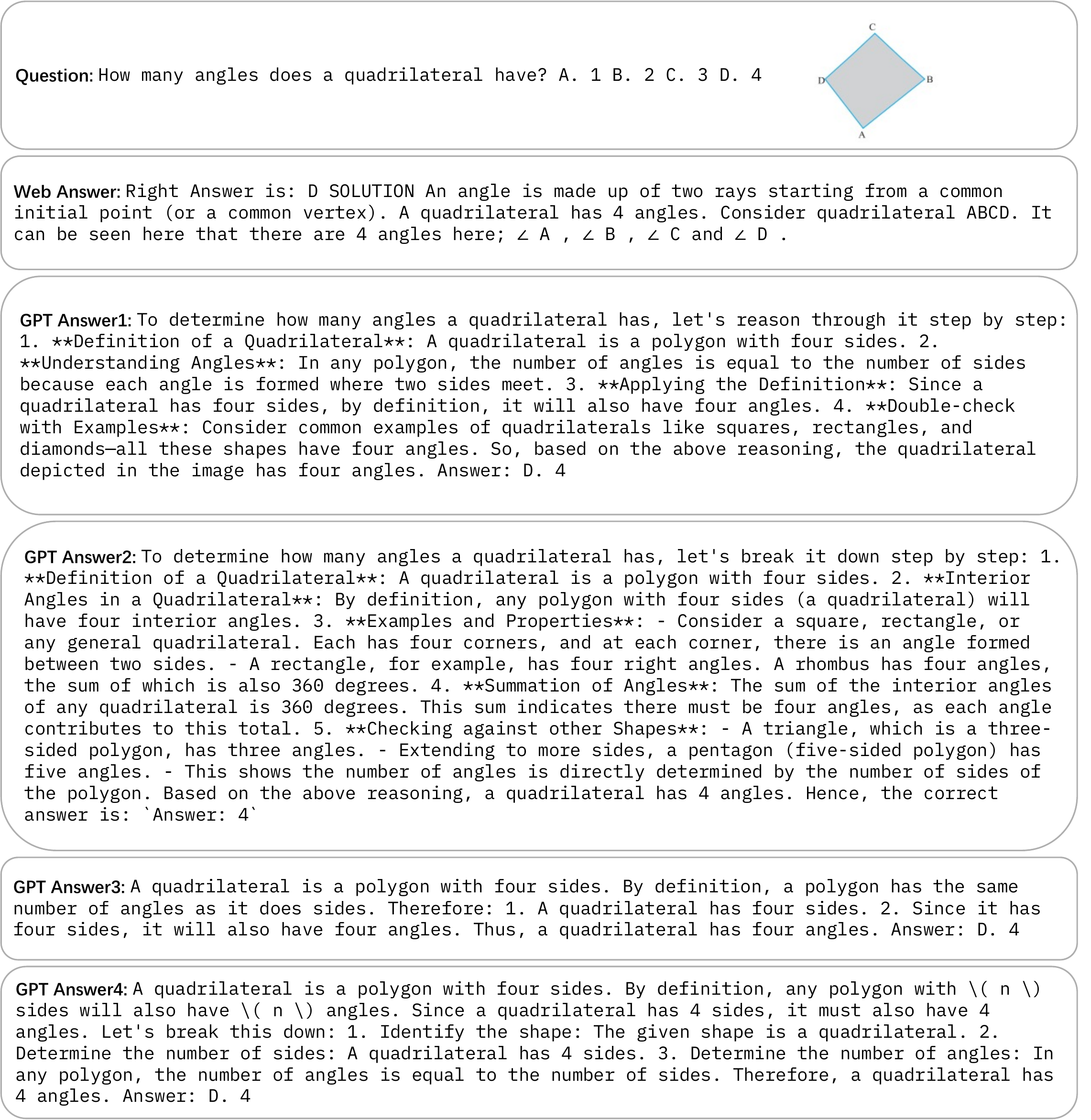}
\end{figure}

\end{document}